\documentclass[letterpaper, 10 pt, conference]{ieeeconf}
\IEEEoverridecommandlockouts
\usepackage{tabularx}
\usepackage{booktabs}
\usepackage{longtable}
\usepackage{multirow}
\usepackage{bbding}
\usepackage[ruled,linesnumbered]{algorithm2e}
\usepackage{graphicx}

\title{\LARGE \bf
	FC-Track: Overlap-Aware Post-Association Correction for Online Multi-Object Tracking
}

\author{Cheng Ju$^{1}$, Zejing Zhao$^{1}$, Akio Namiki$^{1}$%
	\thanks{*This work was not supported by any organization}%
	\thanks{$^{1}$Cheng Ju, Zejing Zhao and Akio Namiki are with the Graduate School of Engineering, Chiba University, Chiba 263-8522 Japan.
		{\tt\small JuchengJC@outlook.com; zejingzh@gmail.com; namiki@faculty.chiba-u.jp}}%
}

\begin{document}

\maketitle
\thispagestyle{empty}
\pagestyle{empty}

\begin{abstract}
Reliable multi-object tracking (MOT) is essential for robotic systems operating in complex and dynamic environments. Despite recent advances in detection and association, online MOT methods remain vulnerable to identity switches caused by frequent occlusions and object overlap, where incorrect associations can propagate over time and degrade tracking reliability.
We present a lightweight post-association correction framework (FC-Track) for online MOT that explicitly targets overlap-induced mismatches during inference. The proposed method suppresses unreliable appearance updates under high-overlap conditions using an Intersection over Area (IoA)–based filtering strategy, and locally corrects detection-to-tracklet mismatches through appearance similarity comparison within overlapped tracklet pairs. By preventing short-term mismatches from propagating, our framework effectively mitigates long-term identity switches without resorting to global optimization or re-identification. The framework operates online without global optimization or re-identification, making it suitable for real-time robotic applications.
We achieve $81.73$ MOTA, $82.81$ IDF1, and $66.95$ HOTA on the MOT17 test set with a running speed of $5.7$ FPS, $77.52$ MOTA, $80.90$ IDF1, and $65.67$ HOTA on the MOT20 test set with a running speed of $0.6$ FPS. Specifically, our framework FC-Track produces only $29.55\%$ long-term identity switches, which is substantially lower than existing online trackers. Meanwhile, our framework maintains state-of-the-art performance on the MOT20 benchmark.

\end{abstract}

\section{INTRODUCTION}
	Recent progress in robotic manipulation has expanded the operational scope of robots from highly controlled environments to more diverse and unstructured real-world scenarios\cite{Oka2017, Sato2020, Zhao2025, Zhao2016, Cao2025}. In working conditions such as logistics automation, healthcare assistance, agricultural operations, and domestic service, robots are required to perceive and interact with multiple surrounding objects in a reliable and timely manner. Effective task execution relies on accurate understanding of both the category and spatial configuration of objects, which directly informs downstream motion planning and control. Vision-based perception, particularly camera sensing, has become a dominant modality for environmental understanding due to its rich information content, flexibility, and ease of deployment. As robots engage in increasingly complex tasks, the ability to simultaneously detect and track multiple objects in real time becomes indispensable. In this context, multi-object tracking (MOT) serves as a core perception component that enables consistent object-level reasoning throughout the task execution process.
	
	MOT aims to detect and identify multiple objects in a video stream or image sequence, which already benefited applications such as video analysis, autonomous vehicles and human activity recognition. Recent years have witnessed rapid advances in MOT methodologies.\cite{Chu2017, Zhang2021, Wang2024, Holz2025} End-to-end methods address MOT problems as objects detection and re-identification (re-ID) in a single network. The inference time is reduced since all features are shared in a single network for both objects detection and re-ID process in this framework. However, suffering from relatively limited tracking robustness, most state-of-the-art methods follow the tracking-by-detection (TBD) paradigm. In this framework, objects are first detected in each frame and then associated with existing tracklets from previous frames or used to initialize new ones. Specifically, association is processed via multiple similarity measures such as Intersection over Union (IoU) and re-ID similarity. 
	
	\begin{figure}[!t]
		\centering
		\includegraphics[width=\linewidth]{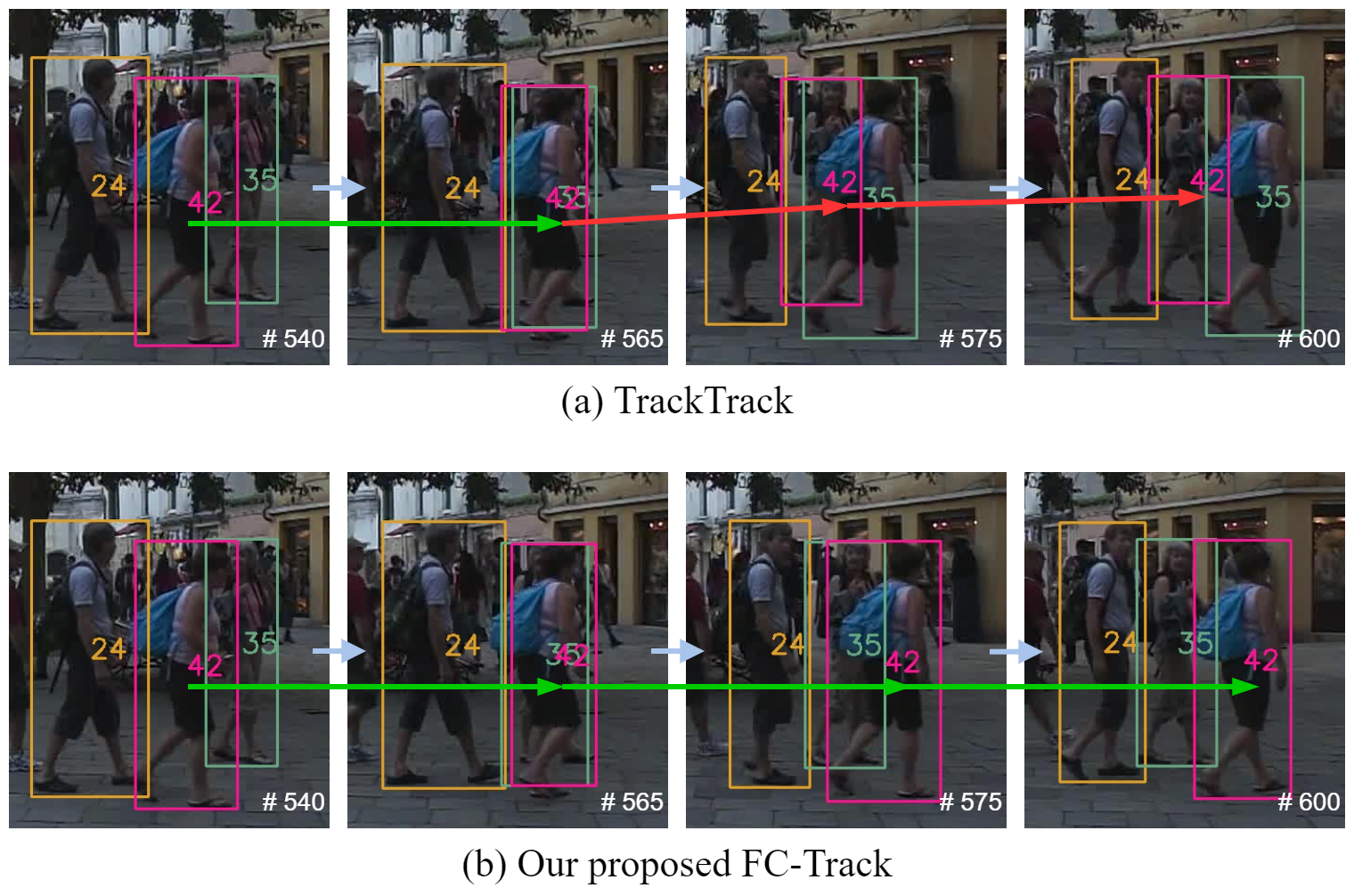}
		\caption{Samples from the results on MOT17 validation set. TrackTrack and our proposed FC-Track use the same detection results. For clarity, we show only a single representative example. On the third frame, TrackTrack exhibits an ID switch when the target overlaps with another target person, our FC-Track maintains consistent tracking throughout the occlusion.}
		\label{fig:Example}
	\end{figure}
	
	Despite significant progress in detection accuracy and association modeling, data association remains a major challenge in practical MOT systems, especially in crowded scenes with frequent occlusions and overlapping objects. Under such conditions, even state-of-the-art trackers inevitably suffer from incorrect matches. More critically, once a mismatched association occurs, the resulting identity error often propagates to subsequent frames, leading to long-term identity switches and severely degraded trajectories. This error accumulation phenomenon reveals a fundamental limitation of many existing online MOT systems to real-world application.
	
	Most existing approaches attempt to mitigate this issue by improving association accuracy through more discriminative appearance representations, sophisticated motion models, or global optimization strategies. While these methods are effective to some extent, these approaches generally treat association decisions as irreversible once made. Although offline or batch optimization methods may revise historical associations, they are typically incompatible with the real-time constraints required in robotics and other online applications. As a result, the problem of online correction of association errors remains underexplored in current MOT pipelines.
	
	In this work, we argue that robust multi-object tracking should not only aim to avoid association errors but also be equipped with the capability to identify and correct mismatches during online inference. Motivated by this insight, we propose a real-time post-association correction framework called FC-Track that continuously refines tracking results after the matching stage. Instead of redesigning the association module, our method acts as a lightweight correction mechanism that detects potential matching errors—particularly those caused by overlapping bounding boxes—and rectifies them before they propagate further in time. This design enables the tracker to recover from short-term mistakes while preserving real-time performance.
	We integrate the proposed framework FC-Track into a standard online MOT pipeline and evaluate its effectiveness on MOT17\cite{Milan2016} and MOT20\cite{Dendorfer2020}. Experimental results demonstrate that our approach improves identity consistency and tracking robustness in challenging scenarios, especially under frequent overlaps and occlusions, with minimal computational overhead. These results suggest that online error correction is a complementary and necessary component for building more reliable MOT systems.
	
	The main contributions of this work can be summarized as follows:
	\begin{enumerate}
		\item We introduce a novel and robust post-association correction framework for online multi-object tracking, which explicitly targets overlap-induced association errors and effectively mitigates long-term identity switches without redesigning the underlying association module.
		\item We propose an overlapping-aware correction mechanism based on the Intersection over Area (IoA) of tracklet bounding boxes and local appearance similarity comparison, enabling reliable identification and reassignment of mismatched detections under severe occlusions and spatial overlap.
		\item We demonstrated the effectiveness and robustness of our method on public dataset MOT17 and MOT20, demonstrating consistent improvements in identity consistency and tracking robustness across diverse crowded-scene scenarios.
	\end{enumerate}

\section{Related Work}
	\subsection{Tracking-by-Detection}
	Tracking-by-detection has become the dominant paradigm in modern multi-object tracking, largely driven by the rapid advances in object detection performance. In this framework, objects are first detected independently in each frame, and temporal consistency is established by associating detections across frames based on motion, appearance, or their combination. 
	For detection part, popular methods such as Faster R-CNN\cite{Ren2015}, SDP\cite{yang2016} and YOLO-based network are widely used. Tracking part is the primary focus for most existing works. SORT\cite{Bewley2016} uses Kalman Filter to track target bounding boxes and uses Hungarian algorithm to associate them. DeepSORT\cite{Wojke2017} extends SORT by incorporating a deep appearance association metric learned offline, enabling more robust data association through occlusions and significantly reducing identity switches. ByteTrack\cite{Zhang2022} improves multi-object tracking by associating both high- and low-confidence detections based on YOLOX detector to reduce missed objects and trajectory fragmentation, achieving state-of-the-art accuracy and robustness across multiple MOT benchmarks. Seg2Track-SAM2\cite{Mendonca2025} integrates pre-trained detectors with SAM2\cite{Ravi2024} and a detector-agnostic Seg2Track module to enable zero-shot multi-object tracking and segmentation, achieving state-of-the-art association accuracy and robust identity management with significantly reduced memory usage. TrackTrack\cite{Shim2025} introduces a track-focused online multi-object tracking framework that employs track-perspective-based association and track-aware initialization to better handle occlusions and assignment ambiguities, achieving superior performance across multiple challenging MOT benchmarks.
	
	\subsection{Occlusion-aware Tracking}
	Occlusion is one of the most persistent challenges in multi-object tracking, often leading to ambiguous associations and identity switches when targets overlap or temporarily disappear. To address this issue, a large body of prior work focuses on incorporating occlusion awareness into the tracking pipeline. SMILEtrack\cite{Wang2024} achieves state-of-the-art multi-object tracking performance by introducing a Siamese network–based similarity learning module with patch self-attention to better distinguish objects with similar appearances, together with a robust tracking framework using a novel GATE function for effective bounding box association. Dong et al.\cite{Dong2017} proposed an occlusion-aware online visual tracking method that mitigates model drift by using a two-stage integrated circulant structure kernel classifier and a classifier pool with entropy-based selection to enable robust and real-time tracking under heavy and long-term occlusions. SiamON\cite{Fan2023} addresses occlusion challenges in visual tracking by introducing a Siamese occlusion-aware network that leverages predefined soft masks and a target-aware attention mechanism to learn robust features under limited occlusion samples, achieving state-of-the-art and real-time performance in occluded scenarios.
	Object overlap, which is more prevalent in crowded scenes, can trigger identity switches that persist over long temporal durations, yet existing tracking pipelines rarely model or correct this error propagation once it occurs.
	
	\subsection{Online Tracking Error Correction}
	While most online MOT methods emphasize reducing the incidence of association errors, some recent approaches explicitly consider correction after errors occur. Unfctrack\cite{Wu2024} mitigates identity switches in multi-object tracking by leveraging unfalsified control to model appearance variation sequences and incorporating a detection–rectification module with an ambiguity-aware association strategy, demonstrating robust performance under occlusions and rapid motion. Zou et al.\cite{Zou2022} proposed a lightweight compensation tracker to alleviate tracking failures caused by missed detections by integrating motion compensation and object selection modules into existing tracking-by-detection frameworks, improving robustness and reducing identity switches without additional networks or retraining. OC-SORT\cite{Cao2023} enhances Kalman filter–based multi-object tracking by adopting an observation-centric strategy that uses detector measurements to construct virtual trajectories during occlusion, effectively correcting error accumulation and improving robustness to occlusion and non-linear motion while maintaining real-time performance.
	In contrast to prior approaches that rely on global re-association to recover from errors, our method performs local and immediate correction within the online tracking loop.
	
\section{Method}
\begin{figure}[thpb]
		\centering
		\includegraphics[width=\linewidth]{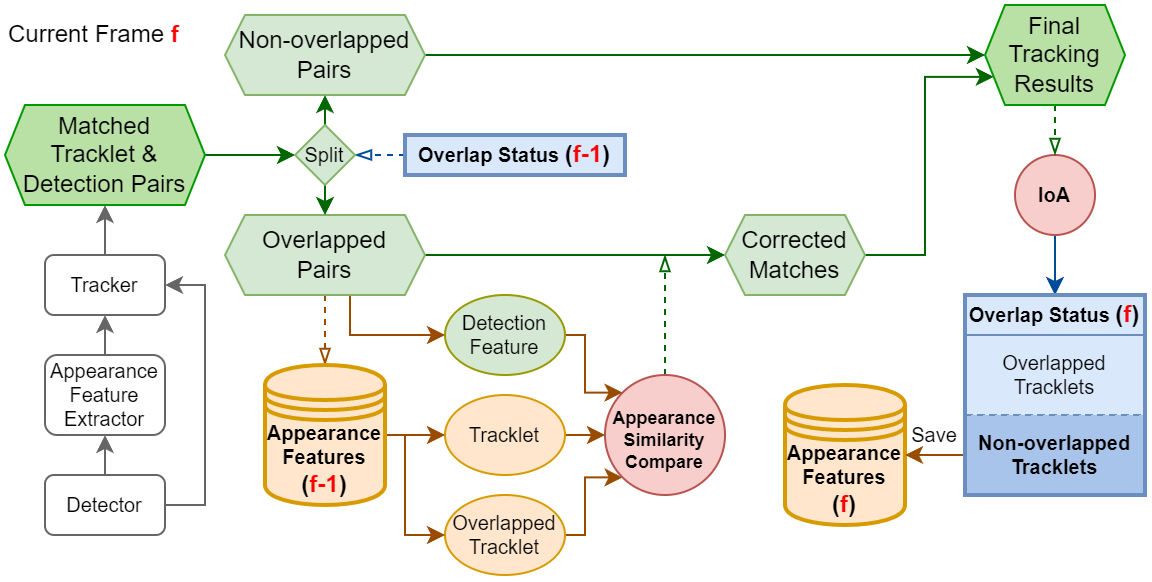}
		\caption{System overview of the proposed post-association correction framework FC-Track. Given matched tracklet–detection pairs at frame $f$, associations are first divided into overlapped and non-overlapped groups using overlap status inherited from frame $f-1$. Non-overlapped pairs are directly accepted, while overlapped pairs are re-evaluated through appearance similarity comparison using stored features and current detections appearance feature. Corrected matches are merged to produce final tracking results. Then the IoA of all tracklets are computed to save as overlap status, and appearance features are updated for the next frame, enabling temporal consistency and online error correction. }
		\label{fig:SystemFig}
	\end{figure}

	We propose a generic and effective association correction framework called FC-Track for online multi-object tracking. Unlike previous works that primarily rely on motion consistency or instantaneous appearance matching, our method explicitly targets association errors caused by heavy spatial overlap, where both motion cues and appearance similarities become unreliable. FC-Track focuses on detecting and correcting the identity switches that arise after the overlapping event to recover the correct identities once the objects become more distinguishable. The proposed correction module operates after the association stage and does not alter the detector or motion model. Whole system design is shown as Fig. \ref{fig:SystemFig}.
	
	Specifically, we first filter the overlapped bounding boxes tracklet pairs in the last frame and save each non-overlapped tracklet appearance features. This step is motivated by that most detection-to-tracklet mismatches happen between heavily overlapped tracking bounding boxes, where which conventional motion-based association usually fails and appearance similarities between competing tracklets are nearly indistinguishable during the overlapping period. 
	Then for each overlapped tracklet pair in the last frame, we compare the appearance similarity distance between initial matched detection and the saved appearance features of both tracklets in overlapped tracklet pair, and reassign detection results to the correct tracklet.
	
	For clarity, we provide the pseudo-code for the whole process in the Algorithm \ref{alg:FCTrack}.
	Detailed descriptions of the overlapping-aware appearance features filtering and mismatches reassignment are provided in the following subsections. To put forwards the state-of-the-art performance of MOT, we imply our association correction method to the high-performance online tracker TrackTrack.
	
	\begin{algorithm}
		\caption{Pseudo-code of FCTrack}
		\label{alg:FCTrack}
			\KwData{
				Current frame $f$; 
				current fram detection $D_{f}$;
				previous frame tracklets $T_{f-1}$; 
				stored tracklets appearance feature $F_{trk}$; 
				detections appearance feature $F_{det}$; 
				overlapped tracklet pairs $P$
			}
			\KwIn{
				detection-to-tracklet matches $M=\left\{(d, t)|d\in D_{f}, t\in T_{f-1}\right\}$;
				update IoA treshold $\tau _{update}$;
				overlap IoA treshold $\tau _{overlap}$;
				minimum correction similarity treshold $\tau _{min}$;
				correction similarity difference treshold $\tau _{dif}$;
			}
			\KwOut{Corrected detection-to-tracklet matches $M_{corr}$}
			\BlankLine
			Initialization: $P\gets \emptyset , F_{det}=\left\{d_{f}.feature|d_{f}\in D_{f}\right\}$\\
			\BlankLine
			$I=[IoA(t_{pri}, t_{aux})|t_{pri}, t_{aux}\in T_{f-1}; pri\neq aux]$\\
			\For{$i$ in $I$}
			{
				\If{$i<\tau _{update}$}
				{
					Update $F_{trk}[t_{pri}]$ from $t_{pri}$\\
					Update $F_{trk}[t_{aux}]$ from $t_{aux}$\\
				}
				\If{$i\geq \tau _{overlap}$}
				{
					$P\gets (t_{pri}, t_{aux})$
				}
			}
			\BlankLine
			\For{$m=(d_{f}, t_{f-1})$ in $M$}
			{
				\If{$t_{f-1}$ in $P.t_{pri}$}
				{
					$(t_{pri}, t_{aux})\gets P[t_{f-1}]$\\
					$S_{pri}\gets Distance(F_{det}[d_{f}], F_{trk}[t_{prime}])$\\
					$S_{aux}\gets Distance(F_{det}[d_{f}], F_{trk}[t_{aux}])$\\
					\If{$S_{pri} >= \tau _{min}$ and $S_{pri} - S_{aux} >= \tau _{dif}$}
					{
						$m\gets (d_{f}, t_{aux})$
					}
				}
				\BlankLine
				\tcc{store new appearance feature}
				\If{$t_{f-1}$ not in $F_{trk}$}
				{
					Create $F_{trk}[t_{f-1}]\gets t_{f-1}$\\
				}
				\BlankLine
				$M_{corr}\gets m$
			}

	\end{algorithm}
	
	\subsection{Overlapping-Aware Appearance Features Filtering}
	To prevent unreliable appearance features caused by severe object overlap from corrupting identity representations, we introduce an overlapping-aware appearance feature filtering strategy that selectively suppresses feature updates and overlapped tracklet pairs under high-overlap conditions on tracklet-level. 
	
\begin{figure}[thpb]
		\centering
		\includegraphics[width=\linewidth]{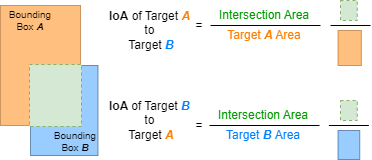}
		\caption{Concept of Intersection over Area (IoA). The overlapped green region represents the intersection between two bounding boxes, while IoA is computed as the ratio of the intersection area to the area of the reference bounding box. Since either box can serve as the reference, two IoA values can be obtained for a box pair.}
		\label{fig:IoA}
	\end{figure}

	At the end of each frame tracking process, we compute the intersection over area (IoA) between all pairs of tracklet bounding boxes to obtain a pairwise relation matrix in which each entry represents the IoA value between two tracklets. Except for the diagonal elements corresponding to self-comparisons, each tracklet forms a one-to-one correspondence with all other tracklets in the matrix. The concept of IoA is shown in Fig \ref{fig:IoA}.
	
	For tracklets which contain IoA values over update IoA threshold $\tau _{update}$, we suspend appearance feature updates and retain the features extracted from the last non-overlapping frame or the initial appearance features, thereby avoiding appearance feature cross-contamination caused by occlusion involved observations. Appearance features of other tracklets are updated and stored normally. 
	When the IoA of one tracklet to another tracklet exceeds the predefined overlap IoA threshold $\tau _{overlap}$, we form an overlapped tracklet pair, where the former tracklet is considered the prime element and the respected latter tracklet is assigned as the auxiliary element. The assignment of prime and auxiliary elements is determined according to the directional IoA relation, where the tracklet whose bounding box area is used as the denominator in the IoA computation is treated as the prime element. This asymmetric definition ensures a consistent and deterministic role assignment for each overlapped pair. This prime–auxiliary relationship is preserved, as the prime tracklet serves as the key to retrieve the initially assigned detection during the subsequent reassignment stage, leveraging the one-to-one correspondence between each matched detection and its associated tracklet.
	For newly initialized tracklets in the current frame, their appearance features are saved as initial identity representations.
	
	\subsection{Mismatches Reassignment}
	To correct detection to tracklet mismatches, we introduce a mismatches reassignment strategy that explicitly targets short-term association errors during online inference. Unlike global re-association or re-identification approaches, our strategy locally corrects detection to tracklet mismatches only under strict conditions. 
	
	For each overlapped tracklet pair in the last frame, we first retrieve the corresponding detection-to-tracklet match from the initial association results of the current frame, using the prime tracklet as the indexing key. Based on this matched detection, we compute two appearance similarity scores: the prime similarity distance, defined as the similarity distance between the detection and the prime tracklet, and the auxiliary similarity distance, defined as the similarity distance between the detection and the auxiliary tracklet within the overlapped pair. If the prime similarity distance is larger than the predefined the minimum correction similarity distance threshold $\tau_{min}$, and the auxiliary similarity distance is smaller than the prime similarity distance and the difference is larger than the similarity distance difference threshold $\tau_{dif}$, we identify this case as a mismatch and reassign the auxiliary tracklet to the detection. Consequently, the prime tracklet is moved to the unmatched tracklet set and the assigned detection is moved to matched detection set, while the auxiliary tracklet is set included in the matched tracklet set and the detection that was originally assigned to the auxiliary tracklet is moved to the unmatched detection set. This design prevents short-term mismatches from propagating into long-duration identity switches.
	
	Considering the growing prevalence of two-stage matching strategies in online MOT, and the two-stage association design of the chosen demonstration tracker, we integrate the proposed mismatch reassignment strategy into each association stage to ensure consistent correction throughout the tracking process.
	
\section{Experiment}
	\subsection{Datasets and Metrics}
	We evaluated our proposed method on 2 datasets: MOT17\cite{Milan2016} and MOT20\cite{Dendorfer2020}. MOT17 training dataset is divided into training half and valid half. MOT17 dataset is used for testing single class multi-objects tracking performance, MOT20 focuses on more complex environments with high-density crowds. The CLEAR\cite{Bernardin2008}, IDF1\cite{Ristani2016} and HOTA\cite{Luiten2021} metrics are used for evaluating overall tracking accuracy.

	\subsection{Implementation Details}
	All evaluating were implemented on single NVIDIA Geforce RTX3090 GPU and Intel Core i9-12900k CPU.
	Except in ablation study, all threshold values are set as follows: the update IoA threshold $\tau_{update}$ to $0.3$, the overlap IoA threshold $\tau_{overlap}$ to $0.8$, the minimum correction similarity distance threshold $\tau_{min}$ to $0.8$ and the similarity distance difference threshold $\tau_{dif}$ to $0.4$, similarity method using cosine distance.
	
	\subsection{Results}
	We compare our FC-Track with existing online state-of-the-art trackers on the MOT17 and MOT20 benchmarks. The results are summarized in Table \ref{tbl:MOT17_SOTA} and Table \ref{tbl:MOT20_SOTA}, respectively. The results of baseline TrackTrack reported in our paper are obtained by running the officially released implementation and submitting the outputs to the MOT benchmark server. These numbers differ slightly from the results reported on the official leaderboard, likely due to differences in implementation versions or evaluation settings.
	
	{\bf MOT17:}
	As shown in Table \ref{tbl:MOT17_SOTA}, Our proposed method FC-Track achieves a HOTA score of $66.95$, which is competitive with current state-of-the-art methods  and improving upon the baseline $66.94$. Our method achieves an IDF1 score of $82.81$, outperforming most existing approaches. In terms of MOTA, our method obtains $81.73$, demonstrating balanced improvements across both aspects. Although the number of ID switches is $837$, which is comparable to other online trackers, our approach maintains strong association accuracy with an AssA of $67.81$. These results indicate that our method achieves balanced performance across detection accuracy and identity preservation.
	
	{\bf MOT20:}
	On the MOT20 dataset, our proposed method FC-Track achieves a HOTA score of $65.67$, outperforming the baseline result of $65.61$. We also observe consistent improvements in IDF1 $80.90$ and competitive MOTA $77.52$, demonstrating robustness in crowded scenes. The ID switch count is $719$, and the AssA score of $67.48$ further indicates reliable association quality under heavy occlusion and frequent interactions.
	
	Although the quantitative results demonstrate competitive performance across standard evaluation metrics, these aggregated scores do not fully reveal how identity errors behave temporally. To better understand this aspect, we further analyze the temporal characteristics of identity switches, focusing on how long trackers remain in an incorrect identity state after a switch. 
	
	\begin{table}[thtp]
		\centering
		\caption{Comparison online results on MOT17 test set.}
		\begin{tabularx}{\linewidth}{lXXXXXX}
			\toprule
			Tracker & HOTA$\uparrow$ & MOTA$\uparrow$ & IDF1$\uparrow$ & AssA$\uparrow$ & IDs$\downarrow$ & FPS$\uparrow$  \\
			\midrule
			MeMOTR & 58.83 & 72.84 & 71.53 & 58.37 & 1902 & 29.6\\
			MOTR & 62.01 & 78.59 & 74.95 & 60.64 & 2619 & 7.5\\
			ByteTrack & 63.05 & 80.25 & 77.30 & 61.98 & 2196 & 29.6\\
			OC-SORT & 63.16 & 78.00 & 77.50 & 63.40 & 1950 & 29.0\\
			BUSCA & 63.92 & 78.63 & 79.20 & 64.25 & 1425 & 35.7 \\
			UTM & 64.04 & \bf 81.78 & 78.70 & 62.54 & 1431 & 13.1 \\
			Deep OC-SORT & 64.88 & 79.37 & 80.58 & 65.93 & 1023 & 28.1 \\
			BoT-SORT & 65.05 & 80.55 & 80.23 & 65.49 & 1212 & 6.8 \\
			MotionTrack & 65.09 & 81.11 & 80.09 & 65.10 & 1140 & 15.7 \\
			RBO-TRACK & 65.25 & 79.49 & 81.89 & 66.42 & 1881 & 16 \\
			UCMCTrack & 65.73 & 80.62 & 80.95 & 66.42 & 1689 & \bf 157.1\\
			TrackTrack & 66.94 & 81.71 & 82.78 & 66.80 & \bf 837 & 5.9 \\
			\bf FC-Track(Ours) & \bf 66.95 & 81.73 & \bf 82.81 & \bf 67.81 & \bf 837 & 5.7 \\
			\bottomrule
		\end{tabularx}
		\label{tbl:MOT17_SOTA}
	\end{table}

	\begin{table}[thtp]
		\centering
		\caption{Comparison online results on MOT20 test set.}
		\begin{tabularx}{\linewidth}{lXXXXXX}
			\toprule
			Tracker & HOTA$\uparrow$ & MOTA$\uparrow$ & IDF1$\uparrow$ & AssA$\uparrow$ & IDs$\downarrow$ & FPS$\uparrow$  \\
			\midrule
			ByteTrack & 61.34 & 77.76 & 75.21 & 59.56 & 1223 & 17.5\\
			OC-SORT & 62.36 & 75.67 & 76.32 & 62.47 & 942 & 5.1\\
			UTM & 62.47 & \bf 78.22 & 76.86 & 61.41 & 1228 & 6.2\\
			UCMCTrack & 62.81 & 75.61 & 77.38 & 63.46 & 1335 & \bf 44.8\\
			MotionTrack & 62.77 & 77.98 & 76.50 & 61.81 & 1165 & 9.0\\
			FineTrack & 63.93 & 76.96 & 78.86 & 64.76 & 1142 & 9.0\\
			SUSHI & 64.33 & 74.29 & 79.80 & 67.47 & \bf 706 & 5.3\\
			TrackTrack & 65.61 & 77.52 & 80.82 & 67.35 & 719 & 0.7\\
			\bf FC-Track(Ours) & \bf 65.67 & 77.52 & \bf 80.90 & \bf 67.48 & 719 & 0.6 \\
			\bottomrule
		\end{tabularx}
		\label{tbl:MOT20_SOTA}
	\end{table}
	
	\subsection{Analysis of ID Switch Duration}
	While ID switches are commonly used to evaluate identity stability, they do not reflect how long an identity error persists once it occurs. In this experiment, we additionally analyze the temporal duration of ID switches in existing multi-object trackers to study whether identity errors tend to be short-lived or prolonged.
	
	For each switch event, we measure its duration, defined as the number of consecutive frames during which a tracker remains associated with an incorrect identity before recovering or terminating. Based on these durations, we compute four statistics: the total number of switches (Count), the average duration frames upon all switches (Mean), the median duration frames upon all switches (Med.), and the proportion of long-term switches exceeding a fixed threshold $\tau _{long}$ over total switches (Long Ratio). This evaluation is conducted on the validation set of the MOT17 dataset. In this experiment, we set long ratio threshold $\tau _{long}$ to $10$, following common practice for all image sequences in MOT17 recorded at frame rates up to $30$ FPS.
	
	The results are shown in Table \ref{tbl:IDSwitchDuration}. Compared with existing methods, our approach achieves consistently shorter switch durations. Specifically, the mean duration decreases from $33.04$ frames in prior trackers to $18.33$, and the median duration drops to $3.0$, indicating that identity errors are corrected more quickly in typical cases. In addition, the long-switch ratio is reduced to $29.55\%$, noticeable lower than all comparison methods, suggesting that our method effectively suppresses persistent identity drift and mitigates long-tail failure cases.
	
	Although the total number of switches is $308$, which is comparable to or slightly higher than other methods, the substantially reduced mean, median, and long-ratio statistics indicate that these switches tend to be short-lived. In other words, even when identity errors occur, the tracker can promptly recover the correct association instead of remaining in an incorrect state for extended periods.
	
	We further analyze identity-based statistics, including Identity True Positives (IDTP), Identity False Positives (IDFP), and Identity False Negative (IDFN). The results show that our method increases the number of IDTP while reducing IDFP compared with the previous tracking methods. This indicates that more frames are correctly associated with the true identity once the correction mechanism is applied. Meanwhile, the IDFN value remains comparable, suggesting that the improvement mainly comes from more reliable identity assignments rather than changes in detection recall. These observations further support the effectiveness of the proposed correction mechanism in maintaining consistent identity associations over time.
	
	These observations demonstrate that our method improves temporal identity stability rather than merely reducing switch frequency.
	
	\begin{table}[thtp]
		\centering
		\caption{ID switch duration results on MOT17 validation set.}
		\begin{tabularx}{\linewidth}{lXXXXXXX}
			\toprule
			Tracker & Count$\downarrow$ & Mean$\downarrow$ & Med.$\downarrow$ & Long Ratio$\downarrow$ & IDTP$\uparrow$ & IDFP$\downarrow$ & IDFN$\downarrow$\\
			\midrule
			ByteTrack & 201 & 33.04 & 11 & 50.25 & 40434 & 13456 & 6951 \\
			BoT-SORT & \bf 199 & 32.89 & 5 & 38.69 & 41757 & 12133 & \bf 6137\\
			TrackTrack & 236 & 22.88 & 5 & 36.86 & 42144 & 11746 & 6927 \\
			FC-Track & 308 & \bf 18.33 & \bf 3 & \bf 29.55 & \bf 42305 & \bf 11585 & 6843 \\
			\bottomrule
		\end{tabularx}
		\label{tbl:IDSwitchDuration}
	\end{table}
	
	\subsection{Ablation Study}
	We conduct a series of ablation experiments to analyze the contribution of each component and design choice in the proposed framework. All ablations are performed on the MOT17 validation set.
	
	{\bf Similarity:}
	We first evaluate different similarity measures used in the correction, including cosine distance and Euclidean distance. As shown in Table \ref{tbl:Abl_Similarity}, cosine distance achieves a HOTA score of $69.67$ and Euclidean distance obtains $69.48$, both outperforming the baseline result of $69.40$. The results indicate that cosine distance provides more reliable identity discrimination, and the consistent improvements under both similarity metrics further verify the effectiveness of our proposed method.
	
	\begin{table}[thtp]
		\centering
		\caption{Similarity ablation study results on MOT17 validation set.}
		\begin{tabular}{lccccc}
			\toprule
			Similarity & HOTA$\uparrow$ & MOTA$\uparrow$ & IDF1$\uparrow$ & AssA$\uparrow$ & IDs$\downarrow$ \\
			\midrule
			Baseline & 69.40 & 76.57 & 81.86 & 73.57 & 400 \\
			Euclidean Distance & 69.48 & 76.49 & 81.90 & 73.71 & 400 \\
			Cosine Distance & \bf 69.67 & \bf 76.60 & \bf 82.12 & \bf 74.08 & \bf 398 \\
			\bottomrule
		\end{tabular}
		\label{tbl:Abl_Similarity}
	\end{table}
	
	{\bf Matching Stage:}
	We next study where to insert the proposed correction module within the tracking pipeline. The current mainstream paradigm adopts a two-stage matching strategy, which is also used by the baseline method TrackTrack. Therefore, we evaluate the effect of integrating our module into the first matching stage and the second matching stage separately. The results show that applying the module in the first stage improves performance, achieving a HOTA score of $69.67$, while inserting it only in the second stage yields no noticeable change $69.40$. This behavior can be explained by the role of each stage: the first stage handles high-confidence associations where early correction can effectively prevent identity drift, whereas the second stage mainly processes low-confidence or ambiguous matches, where errors are more difficult to rectify and thus the correction module has limited impact.
	
	\begin{table}[thtp]
		\centering
		\caption{Matching stage ablation study results on MOT17 validation set.}
		\begin{tabular}{ccccccc}
			\toprule
			Stage 1 & Stage 2 & HOTA$\uparrow$ & MOTA$\uparrow$ & IDF1$\uparrow$ & AssA$\uparrow$ & IDs$\downarrow$ \\
			\midrule
			\XSolid & \XSolid & 69.40 & 76.57 & 81.86 & 73.57 & 400 \\
			\Checkmark & \XSolid & 69.67 & 76.60 & 82.12 & 74.08 & 398 \\
			\XSolid & \Checkmark & 69.40 & 76.57 & 81.86 & 73.57 & 400 \\
			\Checkmark & \Checkmark & 69.67 & 76.60 & 82.12 & 74.08 & 398 \\
			\bottomrule
		\end{tabular}
		\label{tbl:Abl_MatchStage}
	\end{table}
	
\begin{figure}[thpb]
		\centering
		\includegraphics[width=\linewidth]{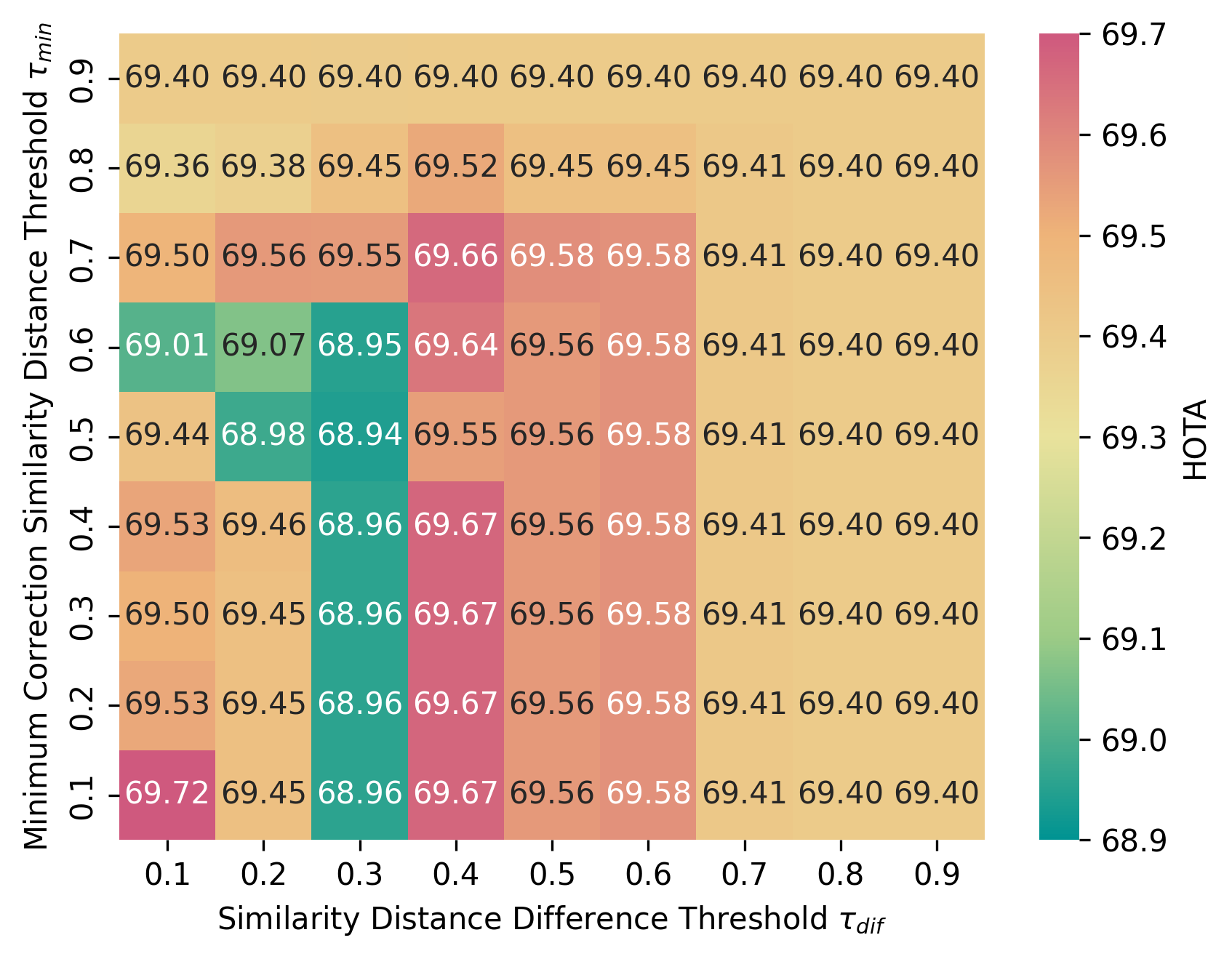}
		\caption{Comparison of the performances of FC-Track over different minimum correction similarity distance threshold $\tau_{min}$ and similarity distance difference threshold $\tau_{dif}$. The results are evaluated on the validation set of MOT17.}
		\label{fig:Abl_SimiThr}
	\end{figure}

\begin{figure}[thpb]
		\centering
		\includegraphics[width=\linewidth]{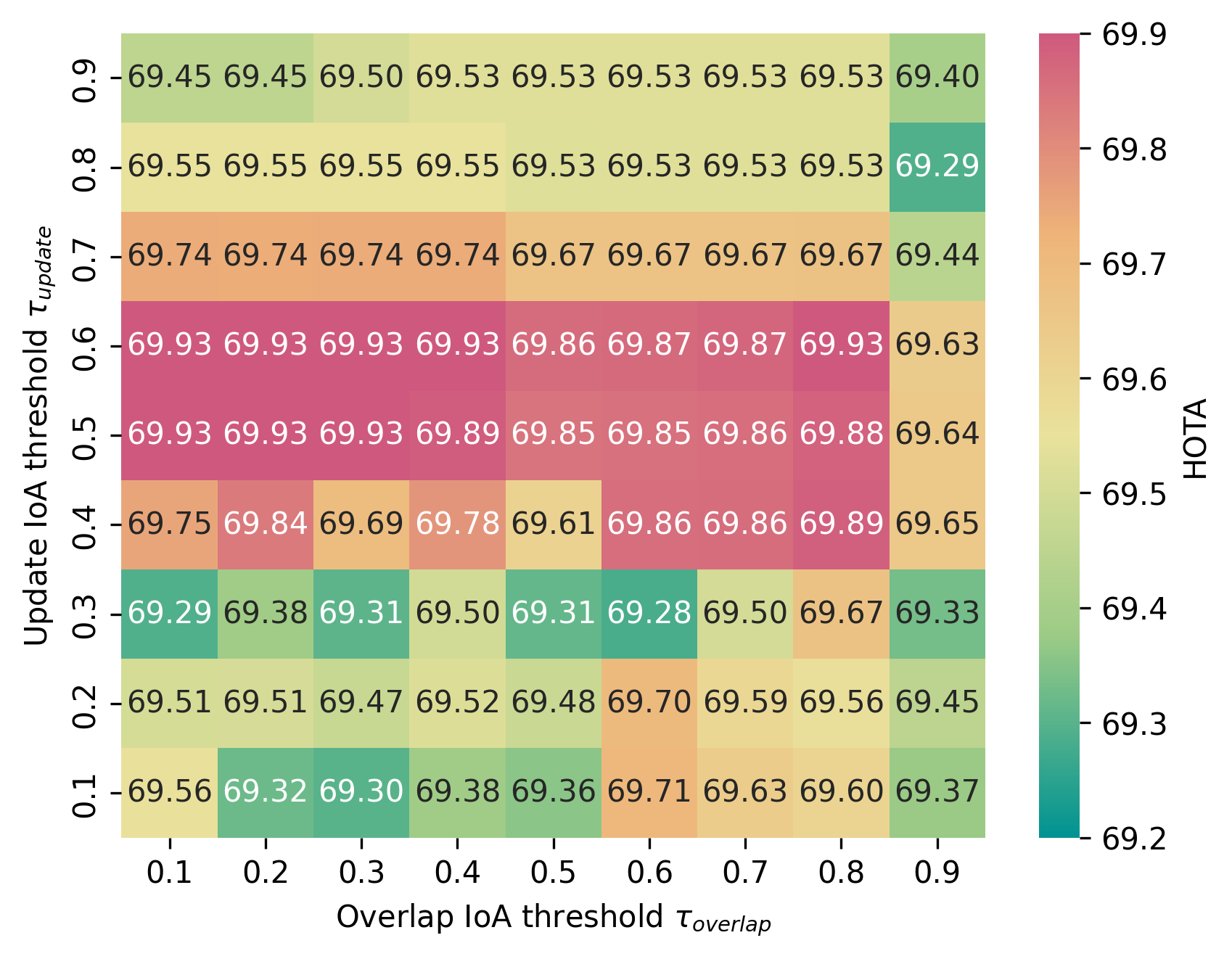}
		\caption{Comparison of the performances of FC-Track over different update IoA threshold $\tau_{update}$ and overlap IoA threshold $\tau_{overlap}$. The results are evaluated on the validation set of MOT17.}
		\label{fig:Abl_IoAThr}
	\end{figure}

	{\bf Threshold Sensitivity:}
	Finally, we analyze the sensitivity of the method to four key thresholds: the update IoA threshold $\tau_{update}$, the overlap IoA threshold $\tau_{overlap}$, the minimum correction similarity distance threshold $\tau_{min}$, and the similarity distance difference threshold $\tau_{dif}$. Two figures are provided for this analysis. The Fig. \ref{fig:Abl_SimiThr} shows the results which set the update IoA threshold $\tau_{update}$ to $0.3$ and the overlap IoA threshold $\tau_{overlap}$ to $0.8$, varies the two similarity thresholds. The Fig. \ref{fig:Abl_IoAThr} shows the results which set the minimum correction similarity distance threshold $\tau_{min}$ to $0.8$ and the similarity distance difference threshold $\tau_{dif}$ to $0.4$, varies the two IoA thresholds. Both figures show that although performance may drop at certain individual threshold values, but the overall trend consistently remains better than the baseline across a wide range of settings, demonstrating the robustness of the proposed method to threshold selection.
	
\section{Conclusion}
	In this paper, we presented a lightweight post-association correction framework FC-Track for online multi-object tracking that explicitly addresses identity switches caused by detection targets overlapping. By suppressing unreliable appearance updates under overlap conditions and introducing a local correction mechanism for mismatched tracklets, the proposed approach effectively prevents short-term association errors from propagating into long-term identity drift. Extensive experiments on MOT17 and MOT20 demonstrate that our method consistently improves holistic tracking quality while maintaining competitive performance across standard metrics and real-time efficiency.
	
	Our method FC-Track first demonstrates clear gains on both MOT17 and MOT20, achieving $81.73$ MOTA, $82.81$ IDF1, and $66.95$ HOTA on the MOT17 test set with a running speed of $5.7$ FPS, $77.52$ MOTA, $80.90$ IDF1, and $65.67$ HOTA on the MOT20 test set with a running speed of $0.6$ FPS. Furthermore, our analysis reveals that the proposed framework significantly shortens the duration of identity switches on MOT17 validation set. Specifically, the mean duration decreases from $22.88$ to $18.33$, the median from $5$ to $3$, and the long-switch ratio from $36.86\%$ to $29.55\%$.
	
	These findings indicate enhanced temporal identity stability rather than merely reducing switch frequency. This property is particularly important for real-world robotic applications, where persistent identity errors can severely affect downstream decision-making. Overall, the results suggest that explicitly modeling overlap-induced ambiguity is a practical and effective direction for improving the robustness of online MOT systems without introducing additional computational overhead or complex global optimization. Furthermore, since the proposed framework operates as a lightweight post-association module, it can be readily integrated with faster or real-time trackers, enabling deployment in real-time robotic applications.

\bibliographystyle{IEEEtran}
\bibliography{references}

	
\end{document}